\documentclass[11pt]{article}
\usepackage{nodalida2025}
\usepackage{times}
\usepackage{url}
\usepackage{latexsym}
\usepackage{comment}
\usepackage{hyperref}
\usepackage{adjustbox}
\usepackage{amsmath}
\usepackage{array}
\usepackage{multirow}

\aclfinalcopy 

\title{Mining for Species, Locations, Habitats, and Ecosystems from \\ Scientific Papers in Invasion Biology: A Large-Scale Exploratory Study with Large Language Models}

\author{
  Jennifer D'Souza$^{1}$, Zachary Laubach$^{2}$, Tarek Al Mustafa$^{3}$, Sina Zarrieß$^{4}$, \\ \textbf{Robert Frühstückl}$^{5}$\textbf{, Phyllis Illari}$^{6}$ \\
  $^{1}$TIB Leibniz Information Centre for Science and Technology, \\
  $^{2}$University of Colorado Boulder, $^{3}$Friedrich Schiller University Jena, $^{4,5}$Bielefeld University, \\ $^{6}$University College London \\    
  {\tt jennifer.dsouza@tib.eu}
}

\date{}

\begin{document}
\maketitle
\begin{abstract}
This paper presents an exploratory study that harnesses the capabilities of large language models (LLMs) to mine key ecological entities from invasion biology literature. Specifically, we focus on extracting species names, their locations, associated habitats, and ecosystems, information that is critical for understanding species spread, predicting future invasions, and informing conservation efforts. Traditional text mining approaches often struggle with the complexity of ecological terminology and the subtle linguistic patterns found in these texts. By applying general-purpose LLMs without domain-specific fine-tuning, we uncover both the promise and limitations of using these models for ecological entity extraction. In doing so, this study lays the groundwork for more advanced, automated knowledge extraction tools that can aid researchers and practitioners in understanding and managing biological invasions.
\end{abstract}


\section{Introduction} 

Human population growth and expansion are coupled with the intentional and unintentional movement of species beyond their historic ranges. 
The introduction of nonnative species can have dramatic impacts that cascade across ecological scales from individual plants and animals to populations and communities \cite{roy2023ipbes}. 
The goal of invasion biology is to identify and understand the impacts of alien species on native flora and fauna across these ecological scales, not only to conserve rare, sensitive, and ecologically valuable native species, but also because intact functional ecosystems provide important economic, public health, and socioemotional services to humans \cite{cassey2018invasion,jeschke2018invasion}. 
Achieving this goal becomes increasingly challenging though, since alien species introductions operate at a rapid rate and global scale, and in the context of increasing human population growth and climate change.
Additionally, the body of research in this domain of ecology is currently growing to an extent that it becomes more and more difficult for invasion biology researchers to build a systematic and shared understanding of the impacts of alien species on ecosystems, or to even comprehensively categorize these species, and their locations, and relationships. This paper explores the potential of recent NLP technologies, such as Information Extraction (IE) approaches based on Large Language Models (LLMs) \cite{llm-amatriain-arxiv,llm-dsouza-orkg}, as a tool that could contribute to predicting future invasions and their consequences.

The extraction and categorization of information from scientific publications is a well-known NLP task \cite{augenstein2017semeval,gabor-etal-2018-semeval,luan2018multi,brack2020domain,dessi2020ai,dsouza-etal-2021-semeval,liu2021uiuc_bionlp,kabongo2021automated,d2022computer,d2024agriculture,shamsabadi-etal-2024-large,d2024overview}. 
In the biomedical domain, NER and RE facilitate large-scale biomedical data analysis, such as network biology \cite{biomed1}, gene prioritization \cite{biomed2}, drug repositioning \cite{biomed3} and the creation of curated databases \cite{biomed4}. 
In the clinical domain, NER and RE can aid in disease and treatment prediction, readmission prediction, deidentification, and patient cohort identification \cite{miotto2018deep}.
However, the domain of ecology and invasion biology in particular has hardly been explored and dedicated datasets are scarce.
To the best of our knowledge, the small-scale INAS dataset \citep{brinner-etal-2022-linking} presents the only invasion biology-specific resource that  provides annotations of hypotheses for scientific abstracts.

This paper presents a study on information extraction (IE) in invasion biology, encompassing both named entity recognition (NER) and relation extraction (RE). We simultaneously build on studies showing that jointly learning NER and RE can enhance overall performance \cite{giorgi2019end} and on recent LLMs which may open new opportunities for IE. 
Thus, our central question is whether LLMs, with their advanced pattern recognition capabilities, can be effectively applied to a new domain to simultaneously identify entities and infer their relationships.
We present an approach that prompts LLMs to identify the following four key entities in invasion biology--viz. species, location, habitat, and ecosystem. Since we cannot evaluate the results quantitatively, we explore them qualitatively with respect to the following questions: (i) do the LLMs extract reasonable species, location, habitat, and ecosystem information and what interactions do they infer? (ii) what relation information is mined?, (iii) what are the benefits of defining LLM workflows to mine large amounts of data? To summarize, this work makes two key contributions: (i) the release of a text data mining corpus of over 10,000 invasion biology papers, including titles, abstracts, full text for nearly 2,000 papers, and structured information extracted by the GPT-4o LLM from the abstracts, to support further research, analysis, and curation (\url{https://doi.org/10.5281/zenodo.13956882}); and (ii) a systematic workflow for schema discovery in information extraction tasks, demonstrated in this paper but broadly applicable for leveraging LLMs in open-ended IE objectives.

\section{Our Text Data Mining Corpus} 

Before tackling the IE task, we compiled a publication corpus as the unstructured source for scientific information. This section outlines the preparation of this dataset for text mining and IE.

\subsection{Defining the Collection}

We started with the Invasion Biology Corpus \cite{mietchen2024invasion}, which lists metadata for 49,438 invasion biology papers in Wikidata. Using the papers' DOIs, we queried the \href{https://ask.orkg.org/}{ORKG ASK} search engine’s \href{https://api.ask.orkg.org/docs}{API} to retrieve abstracts and full texts. ASK was chosen for its convenient access to over eight million publications \cite{knoth2023core} across diverse scientific fields.

\subsection{Corpus Statistics}

Of the 49,438 DOIs queried, 12,636 were available in the ASK database. Among these, 9,802 had abstracts but not full text, and 2,834 included both. This distribution reflects a common issue: open-access full-text articles are limited, posing a challenge for scientific NLP efforts. As a result, our final dataset—consisting of a larger set of abstracts and a smaller set of full texts—serves as the target collection for text mining and IE.

The abstracts range from 10 to 1,608 tokens (average: 235), while the full texts span 28 to 123,958 tokens (average: 7,667), highlighting the greater informational depth of full texts despite their limited availability.

\subsection{Bibliometric Analysis}

\begin{figure}[!htb]
    \centering
    \includegraphics[width=\linewidth]{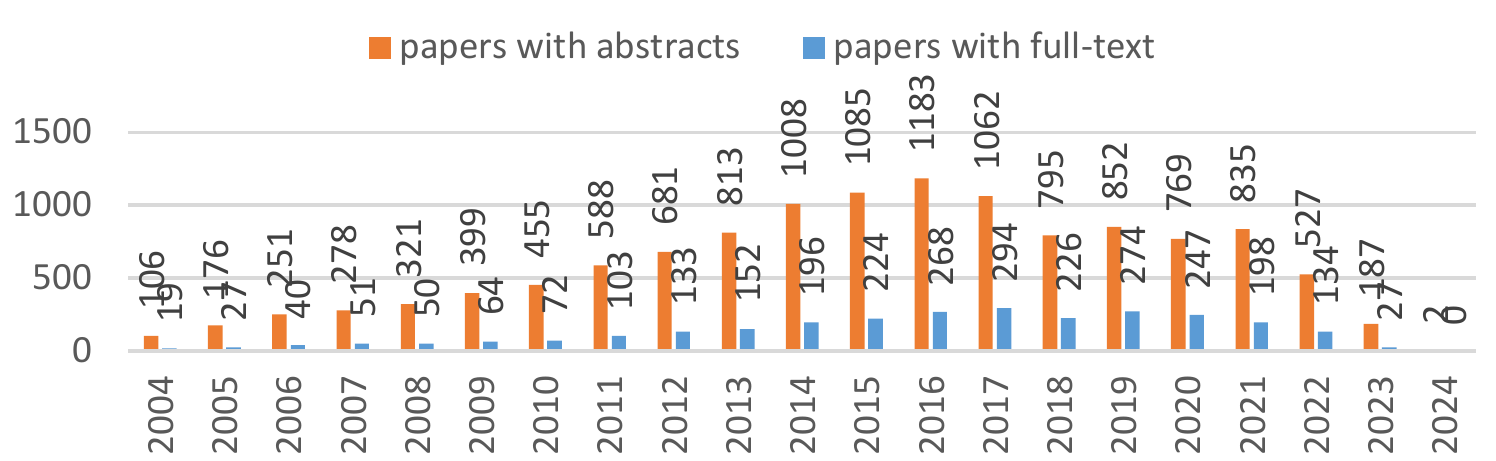}
    \caption{Distribution of papers in our corpus with abstracts and with full text over the past 20 years.}
    \label{fig:bib-stats1}
\end{figure}

\begin{figure}[!htb]
    \centering
    \includegraphics[width=\linewidth]{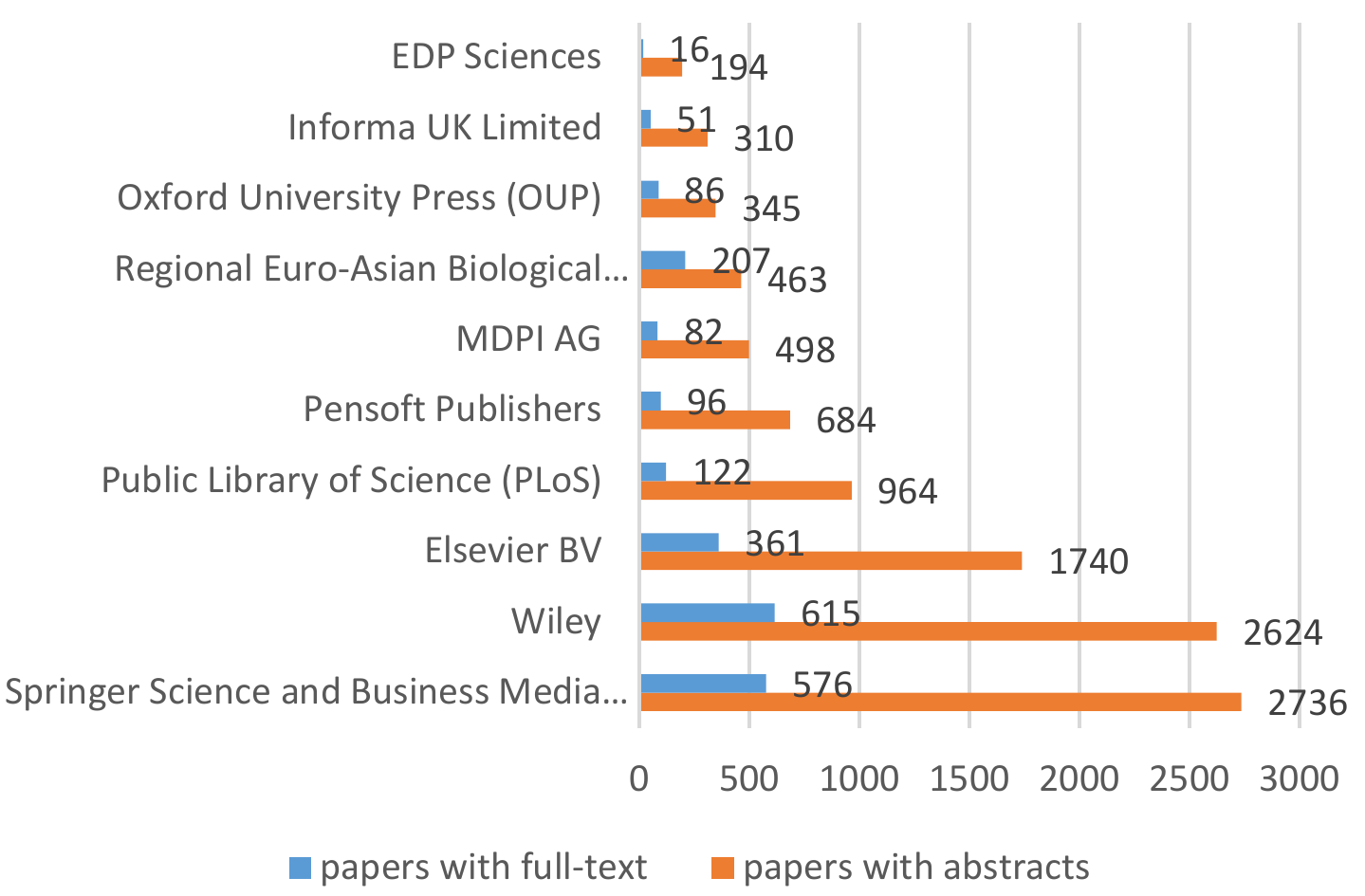}
    \caption{Distribution of papers, in our corpus, by abstract and full-text availability across the top ten publishers.}
    \label{fig:bib-stats2}
\end{figure}

Before the IE tasks, we highlight two key bibliometric insights from our corpus. Among 12,636 papers with abstracts, publication years span 52 years, starting in 1950. Full-text availability begins in 1990. \autoref{fig:bib-stats1} shows a snapshot of the past 20 years, with 2016 having the most abstracts (1,183) and 2017 the most full texts (294). \autoref{fig:bib-stats2} highlights the top ten publishers. Additional insights are available in our \href{https://github.com/jd-coderepos/invasion-biology-IE/blob/main/README.md}{online repository}.


\section{Information Extraction (IE) with Large Language Models (LLMs)} 

An IE task entails two prerequisites. 1) The collection of papers on the which the IE task should be performed. And 2) the schema which defines the IE extraction targets of interest. Having compiled our publication corpus above, we proceed with IE.

\begin{table*}[!htb]
\centering
\begin{tabular}{|l|p{14cm}|}
\hline
\textbf{Entity} & \textbf{Description} \\ \hline
\textbf{Species} & Includes specific named species (e.g., Asterias amurensis) and broader categories (e.g., demersal fish, aquatic invertebrates), covering plants, animals, fungi, or microbes introduced to new environments where they establish, spread, and cause ecological or economic impacts. Higher-level taxonomic or functional groups are included when specific species are not identified, but generic terms like ``invasive species'' are excluded. \\ \hline
\textbf{Location} & Refers to study sites, from specific locations (e.g., ``Port Phillip Bay, southern Australia'') to broader regions (e.g., southern Australia, Amazon rainforest). Includes natural features (rivers, bays, mountains) and administrative areas (cities, states, countries). \\ \hline
\textbf{Ecosystem} & A system of interacting biological and abiotic components, often spanning multiple locations (e.g., the savannah ecosystem across Kenya and Tanzania). \\ \hline
\textbf{Habitat} & A specific part of an ecosystem where an organism lives, such as crocodiles in freshwater habitats (e.g., rivers) within the savannah ecosystem. \\ \hline
\end{tabular}
\caption{\label{table:entities} Definitions of the four entities that encompass the information extraction (IE) aim of this paper.}
\end{table*}

\subsection{Schema Discovery}

Schema or semantic model discovery is a key focus of this subsection. We aim to use LLMs to propose a standard semantic structure for extracting information from each paper. This standardized IE format facilitates downstream processing of the structured data. Since no predefined set of relations exists, the schema must flexibly capture extracted entities and represent their relationships.

We approach the task in two main stages: \textbf{specialize} and \textbf{generalize}. In the \textbf{specialize} stage, the LLM is instructed to generate a schema as the target for IE, focusing on four entities: \textit{species}, \textit{location}, \textit{habitat}, and \textit{ecosystem}. The schema proposed by the LLM includes an IE objective with properties that specify relationships between the extracted entities. This behavior is guided by a system prompt, and the LLM processes each paper instance individually through user prompts to define a semantic model tailored or specialized to each paper. In the \textbf{generalize} stage, the LLM is instructed to generate a generalized schema based on input of the specialized schemas created per paper in the first stage. The generalized schema would then represent the most flexible way to capture the relations between the entities across all papers. The following sections provide detailed descriptions of each stage.

\subsubsection{Stage Specialize: Schemas per Paper}
\label{sec:specialize}

The LLM is the central tool in this endeavor, operating in the text \textit{completion} or \textit{generation} mode. The interaction with the LLM is guided by two types of prompts: the \textsc{system prompt}, which defines the model's persona and specialized behavior for a given task, and the \textsc{user prompt}, which supplies input data for the defined behavior. The LLM’s behavior, once established by the system prompt, remains consistent throughout the interaction session per stage.

\textbf{\textsc{System prompt.}} The system prompt defines the LLM's persona and provides structured guidance for the task. It is composed of three key components: (a) Role, (b) Task instruction, and (c) Output format. In this stage, the LLM assumes the overarching role of a ``research assistant in invasion biology or ecology tasked with reading and understanding scientific papers for extracting relevant information within a schema.'' Its primary instruction is to extract terms related to four entities—\textit{species}, \textit{location}, \textit{habitat}, and \textit{ecosystem}—and identify the relationships expressed in each paper. Additionally, the LLM is directed to structure the extracted information into an external container, representing the schema that defines the extraction target.

To refine the LLM's performance, the system prompt was iteratively improved over two runs. In the first run, the \href{https://github.com/jd-coderepos/invasion-biology-IE/blob/main/LLM-based%20IE/1-specialize/data/data-v1/system-prompt.txt}{prompt} lacked detailed definitions of the entities. After discussions with an ecologist, the second run included precise definitions of the field of invasion biology and the four entities (see \autoref{table:entities}). The \href{https://github.com/jd-coderepos/invasion-biology-IE/blob/main/LLM-based%20IE/1-specialize/data/data-v2/system-prompt.txt}{final system prompt}, incorporates these definitions, enhancing the LLM’s clarity and consistency in identifying and extracting the specified entities and their relationships.

This refinement aligns with the principles of in-context learning \cite{radford2019language}, which emphasize that providing clear and detailed task instructions ensures the model's better comprehension of the downstream task, leading to more accurate and consistent results.\footnote{We acknowledge extensive literature on in-context learning advocating for few-shot task examples \cite{few-shot-learning}. In this paper, we interpret in-context learning as only providing a detailed task description without examples.} By supplying the LLM with a well-defined context, we enable it to successfully extract and organize the desired information from scientific texts.

\textbf{\textsc{User prompt.}} The model is given one paper at a time and is allowed to suggest its own semantic model. The instruction in the user prompt is simple and is as follows: ``Extract the information as instructed from this article title and abstract. \text{\textbackslash n\textbackslash n} \textbf{Title:} \{title\}\text{\textbackslash n} 
\textbf{Abstract:} \{abstract\}\text{\textbackslash n}''.

\paragraph{Result.} Ten papers were randomly selected, and the LLM was prompted to generate a schema for each, which was stored and is available in the data folder of our \href{https://github.com/jd-coderepos/invasion-biology-IE/blob/main/LLM-based%20IE/1-specialize/README.md}{online repository}. The tenth paper, an outlier from the Wikidata Invasion Biology dataset, highlights the potential presence of false positives—papers unrelated to invasion biology.

The schemas evolve significantly to better capture relationships across the nine true-positive papers. Early schemas, like \href{https://github.com/jd-coderepos/invasion-biology-IE/blob/main/LLM-based%20IE/1-specialize/data/data-v2/schema-paper1.txt}{Schema 1}, use basic categorizations of species, locations, and ecosystems with simple binary relationships. In contrast, later schemas, such as \href{https://github.com/jd-coderepos/invasion-biology-IE/blob/main/LLM-based%20IE/1-specialize/data/data-v2/schema-paper8.txt}{Schema 8} and \href{https://github.com/jd-coderepos/invasion-biology-IE/blob/main/LLM-based%20IE/1-specialize/data/data-v2/schema-paper9.txt}{Schema 9}, adopt more sophisticated structures, accommodating complex interactions involving multiple entities and contextual factors. E.g., relationships like ``introduced\_species\_location'' or ``ecological\_disturbance'' incorporate broader ecological and anthropogenic processes, moving beyond pairwise connections. This progression enhances granularity, capturing nuanced interactions like species dynamics influenced by environmental conditions or human activity, making the schemas more meaningful for analysis and synthesis.

The schemas reveal recurring patterns and unique adaptations in representing ecological data. Common themes, such as invasion biology, pollination networks, and anthropogenic impacts, highlight key areas of focus. Standardized fields like \textit{species} and \textit{location} ensure consistency, while tailored relationship labels, such as ``most effective pollinators'' (\href{https://github.com/jd-coderepos/invasion-biology-IE/blob/main/LLM-based%20IE/1-specialize/data/data-v2/schema-paper2.txt}{Schema 2}) and ``competitive replacement'' (\href{https://github.com/jd-coderepos/invasion-biology-IE/blob/main/LLM-based%20IE/1-specialize/data/data-v2/schema-paper5.txt}{Schema 5}), add specificity and contextual relevance. The inclusion of spatial and environmental parameters, like \textit{habitats} and \textit{ecosystems}, emphasizes their role in shaping ecological interactions. These schemas balance standardization and flexibility, illustrating how modular design can synthesize diverse unstructured text data while preserving study-specific distinctions.

\begin{table*}[h!]
\centering
\begin{tabular}{|p{0.15\textwidth}|p{0.25\textwidth}|p{0.5\textwidth}|}
\hline
\textbf{Extraction Target} & \textbf{Extracted Item} & \textbf{Extracted Item Properties} \\
\hline
\textbf{Species} & \multirow{2}{0.25\textwidth}{\texttt{name}: species\_name} & \texttt{role}: native/introduced/alien/invasive \\
                  & & \texttt{taxonomy\_level}: species/genus/family \\
\hline
\textbf{Location} & \multirow{3}{0.25\textwidth}{\texttt{name}: location\_name} & \texttt{category}: natural/administrative \\
                   & & \texttt{geopolitical\_info}: country/region/city \\
                   & & \texttt{additional\_details}: climatic/physiographic \\
\hline
\textbf{Ecosystem} & \multirow{2}{0.25\textwidth}{\texttt{name}: ecosystem\_name} & \texttt{type}: aquatic/terrestrial/marine \\
                    & & \texttt{scope}: local/regional/global \\
\hline
\textbf{Habitat} & \multirow{3}{0.25\textwidth}{\texttt{name}: habitat\_name} & \texttt{type}: aquatic/terrestrial/marine \\
                  & & \texttt{subcomponent\_of}: ecosystem\_name \\
                  & & \texttt{specifics}: e.g., benthic, litoral \\
\hline
\textbf{Relationships} & \multirow{4}{0.25\textwidth}{\texttt{related\_entities}: [entity1, entity2, ...]} & \texttt{name}: relationship\_name \\
                        & & \texttt{type}: biological/physical/ecological/anthropogenic \\
                        & & \texttt{directionality}: unidirectional/bidirectional \\
                        & & \texttt{context}: relationship\_contextual\_description \\
\hline
\end{tabular}
\caption{Standardized information extraction (IE) schema for four ecological entities, their relationships, and associated properties, pertinent to structure information from invasion biology scientific papers.}
\label{tab:final_schema}
\end{table*}

\subsubsection{Stage Generalize: Generic Schema}
\label{sec:generalize}

In this stage, the goal was to develop a standardized container for the extracted information in the form of a JSON data structure, specifically designed to capture the relationships between four entities: species, location, habitat, and ecosystem. The system prompt given to the LLM was similarly structured as the \textbf{specialize} stage, with the model’s role defined not only as a research assistant in invasion biology but also as having expertise in semantic modeling. Overall, the prompt consisted of the same structural components: (a) Role, (b) Task instruction, and (c) Output format. However, aside from the role, the task instruction defining the LLM’s behavior in this stage was different. At this stage, the LLM was expected to review each individual schema from the earlier stage for every paper and then thoughtfully propose a standardized schema. This is related to prior work \cite{baazizi2017schema,baazizi2020human} which presents an approach based on discovering a schema for individual JSON documents and then merging each of these schemas. In our case, we discover schemas from the unstructured texts of individual papers and then merge them in this stage. This schema would eventually form the IE objective, to be applied across all papers in the collection to extract the relevant entities and their relationships.

Since LLMs are known to generate different responses to the same prompt, the LLM was prompted thrice with the same task each time prompting the model in the same way: ``Read the nine different schema instances and generate a standardized schema in the JSON output format.'' 

\paragraph{Result.} The three JSON schema variants generated represented a thoughtful approach to standardizing the representation of entities and their relationships in invasion biology. The common thread across the 3 standardized schema responses was designed to capture information about four primary entities—species, location, habitat, and ecosystem—along with their properties and interrelations, enabling consistent and structured extraction of relevant data. The standardized \href{https://github.com/jd-coderepos/invasion-biology-IE/blob/main/LLM-based%20IE/2-generalize/data/schema-variant1.txt}{schema response 1} emphasized geospatial and hierarchical context, incorporating coordinates for locations and explicitly linking habitats to ecosystems. \href{https://github.com/jd-coderepos/invasion-biology-IE/blob/main/LLM-based%20IE/2-generalize/data/schema-variant2.txt}{Schema 2} introduced detailed descriptions and a focus on the roles of species (e.g., native, introduced, invasive), while expanding relationship types to include biological, physical, and anthropogenic interactions. The standardized \href{https://github.com/jd-coderepos/invasion-biology-IE/blob/main/LLM-based%20IE/2-generalize/data/schema-variant3.txt}{schema response 3} provided rich categorization, including species taxonomy and physiographic attributes for locations, along with specific habitat details such as benthic and littoral zones. Together, these schemas, while sharing commonalities, reflected minor differences in terms of the respective knowledge capture priorities, from geospatial precision to descriptive and taxonomic richness.

Based on these observations, we finalized the \href{https://github.com/jd-coderepos/invasion-biology-IE/blob/main/LLM-based%20IE/2-generalize/data/schema-finalized.json}{schema}, which organizes data around core entities—species, locations, ecosystems, habitats, and relationships—each with structured properties tailored to ecological contexts. For example, species are characterized by roles (e.g., native, invasive) and taxonomy levels (e.g., genus, family), while locations are enriched with geopolitical and physiographic details. Ecosystems and habitats are distinguished by type and scope, with hierarchical links between habitats and ecosystems enhancing clarity. Relationships capture interactions between entities, defined by type (e.g., biological, ecological) and directionality. This schema enables precise mapping of ecological networks and differentiation of key factors, such as invasive versus native species, across datasets. A detailed breakdown of the schema is presented in \autoref{tab:final_schema}.

\subsection{Information Extraction}
\label{sec:IE}

With a standardized semantic structure for extracting information from each paper, enabling easier downstream processing, the LLM-based IE task was conducted.

\subsubsection{Stage Extract: Populate Schema}
\label{sec:extract}

This stage now fulfills the main objective of this work, i.e. to extract information from a large-scale corpus (12,636 in our case) with an LLM to mine species, location, habitat, and ecosystem entities and their relations. The system prompt in this stage was close to the \textbf{specialize} stage system prompt where the role specified for the LLM was ``research assistant in invasion biology or ecology tasked with reading and understanding scientific papers \textit{to} extract relevant information \textit{per the given predefined schema}.'' 

\subsection{Technical Details}

The proprietary OpenAI \href{https://openai.com/index/hello-gpt-4o/}{GPT-4o model} was used for all tasks in this paper. Schema generation tasks in the \textbf{specialize} (Section \ref{sec:specialize}) and \textbf{generalize} (Section \ref{sec:generalize}) stages were completed in a few seconds per schema. For the full extraction task in the \textbf{extract} stage (Section \ref{sec:extract}), which applied the extraction schema to a corpus of 12,636 papers, the LLM took approximately three days. 

The resulting dataset is available on \href{https://doi.org/10.5281/zenodo.13956882}{Zenodo}. The code for running the GPT-4o model is accessible in our \href{https://github.com/jd-coderepos/invasion-biology-IE/blob/main/LLM-based%20IE/3-extract/README.md}{GitHub repository} and supports processing either a \href{https://github.com/jd-coderepos/invasion-biology-IE/blob/main/LLM-based%20IE/3-extract/code/gpt-extract.py}{single paper} or a \href{https://github.com/jd-coderepos/invasion-biology-IE/blob/main/LLM-based%20IE/3-extract/code/gpt-bulk-extract.py}{collection of papers}.

\subsection{Results and Discussion}

Of the 12,636 papers in our dataset, the LLM classified 1,740 as outside the scope of invasion biology, responding with ``N/A'' and skipping extraction for these papers. This left 10,896 papers as the target for IE. This section summarizes the results.

The first exploratory insight highlights the diverse roles of species extracted by the LLM, reflecting their origins, behaviors, ecological functions, and impacts within invasion biology. Broad categories include \textbf{native}, \textbf{alien}, \textbf{introduced}, \textbf{invasive}, and \textbf{naturalized}, alongside more specific roles such as \textbf{agricultural weeds}, \textbf{biological control agents}, \textbf{pathogens}, \textbf{mutualists}, and \textbf{ecosystem engineers}. Some roles emphasize species’ origins (e.g., \textbf{indigenous}, \textbf{non-native}, \textbf{cryptogenic}), while others highlight their behaviors (e.g., \textbf{invasive}, \textbf{colonizer}, \textbf{expanding}) or ecological functions (e.g., \textbf{symbiont}, \textbf{facilitator}, \textbf{pioneer}). Additionally, certain roles capture species’ interactions within ecosystems, such as \textbf{co-introduced species}, \textbf{specialist herbivores}, or \textbf{cryptic invaders}, while others underline their relevance to conservation and management, such as \textbf{natural enemies}, \textbf{candidate biological control agents}, or \textbf{quarantine pests}. Together, this spectrum of roles underscores the complexity of species dynamics in invasion biology, providing valuable insights into biodiversity patterns, ecosystem impacts, and strategies for conservation and management. A detailed list of all roles is available \href{https://github.com/jd-coderepos/invasion-biology-IE/blob/main/LLM-based%20IE/3-extract/analysis/unique-roles-observed.txt}{here}.

We conducted a finer-grained exploration of the extracted data, focusing on the most prevalent species names within the roles of \textit{invasive}, \textit{native}, and \textit{introduced} species. \textbf{Invasive species} dominated, with examples such as \textit{Procambarus clarkii} (76 mentions), \textit{Harmonia axyridis} (73 mentions), and \textit{Rhinella marina} (68 mentions) highlighting well-documented invaders. \textbf{Native species}, though less frequent, included specific examples like \textit{Austropotamobius pallipes} and \textit{Phragmites australis} (24 mentions each). \textbf{Introduced species}, while fewer, included \textit{Oncorhynchus mykiss} and \textit{Crassostrea gigas}, reflecting their dual perspectives of their ecological integration and potential invasiveness. Overall, the data underscores the breadth of roles species play in invasion biology, from invaders disrupting ecosystems to native species requiring conservation focus and introduced species with varying impacts. However, the extraction also included generic terms (e.g., ``native species'' and ``native plants'') as species names, indicating noise from the unsupervised nature of the IE task and highlighting the need for post-filtering to refine the analysis. The full list is available \href{https://github.com/jd-coderepos/invasion-biology-IE/blob/main/LLM-based%20IE/3-extract/analysis/species-role-counts.csv}{here}.

The dataset highlights key \textbf{geopolitical locations}, with the five most frequent countries being Australia (406), South Africa (248), New Zealand (236), Italy (187), and France (168). Prominent regions include Europe (601), North America (348), the Mediterranean Sea (117), Asia (112), South America (98), the Mediterranean (80), and the Iberian Peninsula (75). Less common mentions were cities like Sydney (8), Hong Kong (7), Rome (6), Cape Town (6), and Brisbane (5). The strong representation of Europe and North America reflects their prominence in the data, while the frequent mentions of Australia, South Africa, and New Zealand suggest a focus on biodiversity hotspots or ecological studies. Overall, the dataset spans a diverse range of geographical entities, from continents and regions to countries and cities, emphasizing a global perspective with notable references to the Mediterranean, Asia, and locations such as California and Queensland. Browse the data used for analysis in this paragraph \href{https://github.com/jd-coderepos/invasion-biology-IE/blob/main/LLM-based%20IE/3-extract/analysis/location-geoinfo-counts.csv}{here}.

Furthermore, the extracted information provides a comprehensive view of \textbf{terrestrial, marine, and aquatic ecosystems}, reflecting their ecological and scientific significance. Terrestrial ecosystems (93) are the most frequently mentioned, with grasslands (42), forests (45), and agricultural landscapes (47) prominently featured, emphasizing biodiversity and land use. Mediterranean ecosystems (37) and tropical ecosystems (26) highlight climate-specific regions, while environments such as riparian (14), alpine (8), and sub-Antarctic (8) systems underscore specialized ecological contexts. Urban ecosystems (46) also feature prominently, reflecting the interplay between natural and human-modified systems. Among marine ecosystems, the Mediterranean Sea (71) is a standout, alongside coral reefs (8), seagrass meadows (3), and regions like the Baltic Sea (12), showcasing marine diversity. Aquatic ecosystems, particularly freshwater systems (199), are strongly represented, with mentions of wetlands (40), lake ecosystems (59), and riverine ecosystems (36), as well as transitional zones like estuarine ecosystems (35) and coastal wetlands (10), which bridge freshwater and marine environments. Collectively, these ecosystems highlight the dataset’s holistic representation of ecological diversity. The full list of the data used for this analysis is \href{https://github.com/jd-coderepos/invasion-biology-IE/blob/main/LLM-based%20IE/3-extract/analysis/ecosystems-type-counts.csv}{here}.

Regarding \textbf{habitat as a subcomponent of ecosystem} extraction, the following insights were obtained: The dataset reveals nuanced relationships where habitat names and ecosystem types differ across aquatic, marine, and terrestrial environments, showcasing ecological diversity and complexity. In aquatic systems, examples include the \textit{pelagic zone} linked to the lake ecosystem, \textit{ballast water} associated with the marine ecosystem, and \textit{estuarine habitat} tied to the estuarine ecosystem, while unique pairings such as \textit{wetland habitat} with wetland ecosystem and \textit{riverine habitat} with freshwater ecosystem further highlight habitat specificity. In marine environments, \textit{kelp beds} are tied to rocky subtidal ecosystems, and \textit{mussel beds} are associated with the ``intertidal zone,'' emphasizing distinct ecological contributions. Human-modified habitats, such as \textit{artificial coastal defense structures} linked to \textit{Biogenic reefs}, underscore the role of anthropogenic features, while connections like \textit{maerl habitat} and \textit{seaweed canopies} to coastal ecosystems highlight the contributions of unique marine features. Terrestrial systems also show diverse relationships, including \textit{forest habitat} tied to forest ecosystem, \textit{soybean fields} linked to agricultural ecosystem, and \textit{forest understory} associated with ``deciduous forest,'' reflecting agricultural and forest-specific ecological roles. Unique pairings, such as \textit{sandy coastal habitats} with ``coastal dunes'' and \textit{orchards} with orchard ecosystem, emphasize distinct features, while \textit{urban areas} and \textit{domestic gardens} tied to urban ecosystems reflect the ecological importance of human-modified environments. Collectively, this illustrates the dataset's detailed representation of \href{https://github.com/jd-coderepos/invasion-biology-IE/blob/main/LLM-based%20IE/3-extract/analysis/habitats-type-counts.csv}{habitat-ecosystem relationships}, highlighting their interconnectedness and ecological significance in diverse environments.

The corpus of invasion biology papers reveals a rich diversity of \href{https://github.com/jd-coderepos/invasion-biology-IE/blob/main/LLM-based%20IE/3-extract/analysis/unique-relation-types-observed.txt}{relation types}, reflecting the interdisciplinary nature of the field. Predominant categories include \textbf{ecological} relations, such as \textbf{ecological/anthropogenic}, \textbf{ecological/economic}, and \textbf{socio-ecological}, which highlight the interactions between natural systems and human influences. Among these, ecological relations are the most prevalent, with \textbf{invasion} (814) dominating, followed by \textbf{competition} (429), \textbf{impact} (349), and \textbf{predation} (301), underscoring key processes driving species interactions and environmental changes. Additional relations like \textbf{colonization} (179), \textbf{distribution} (179), and \textbf{habitat preference} (123) emphasize the dynamics of species spread, habitat utilization, and behavioral patterns. Complementing this, \textbf{biological relations} such as \textbf{predation} (177), \textbf{parasitism} (151), and \textbf{hybridization} (74) focus on specific ecological interactions, while associations like \textbf{pollination} (25) and \textbf{symbiosis} (25) highlight cooperative and movement-based dynamics. Furthermore, \textbf{physical relations}, including \textbf{location}, \textbf{transport}, and \textbf{introduction location}, emphasize spatial and movement dynamics critical to understanding alien species impacts and habitat transformations. Lastly, \textbf{anthropogenic relations} are dominated by \textbf{introduction} (157), along with \textbf{introduction pathway} (45) and \textbf{transportation} (14), underscoring the significant role of human activities in species dispersal and ecological consequences. Collectively, these relations capture the multifaceted processes and interactions that define invasion biology.

The wealth of information mined from a fully unsupervised IE task conducted by an LLM highlights the immense potential of large language models as tools for researchers and practitioners in ecological sciences and beyond. These models have the potential to serve as valuable assistants in tasks such as systematic or scoping reviews. The results presented here represent only a fraction of the insights that can be derived from the extracted information in our large corpus of over 10,000 papers. To support further research, we have released the corpus at \url{https://doi.org/10.5281/zenodo.13956882}.

This work aligns with the paradigm of open information extraction (OIE) \cite{etzioni2008open,fader2011identifying,etzioni2011open}, which traditionally relies on syntactic patterns in text, such as verbs for relations and (subject, object) dependency structures. However, the advanced semantic comprehension capabilities of LLMs have far surpassed traditional OIE methods. By extending beyond syntax into the realm of semantics, LLMs provide highly practical solutions for analyzing complex relationships in large-scale corpora.

\section{Recommendations for Future Work}

The semantic web community presents technologies and concepts that may supply direction to the future work on this topic. Ontologies have been a focal point of research in the domain for decades, however, the expertise required for their use has presented a barrier to outside establishment. With the rise in prominence of LLMs in both research and application, a new branch of research has emerged with the goal to investigate the interplay of ontologies and LLMs for more precise IE and to facilitate creating linked data. This branch currently covers the following topics: How can LLMs support ontology and knowledge graph construction? \cite{kommineni2024human} How can LLMs for question answering be improved through ontology support? \cite{allemang2024increasing} How can LLMs support ontology learning from texts? \cite{babaei2023llms4ol,llms4ol2024} How can representation learning of LLMs be enhanced by ontologies? \cite{ronzano2024towards}


The concept of ontologies, defined as \textit{a formal, explicit specification of a shared conceptualisation} \cite{studer1998knowledge}, highlights their role in enabling \textit{a shared understanding of a domain} and reflecting \textit{consensus about domain knowledge}. This study builds on these principles, exploring how information can be extracted from unstructured texts and structured using semantic models. Future work can extend this approach by integrating ontological knowledge through schema-driven IE, addressing key questions: What information should be provided to LLMs to improve extraction, and where in the workflow is it most effective? How can LLMs be precisely guided \cite{caufield2024structured} toward achieving semantic modeling objectives? Finally, how does shared human consensus in a domain align with the consensus extracted by LLMs trained on human knowledge? Exploring these questions in future work can bridge the gap between human and machine understanding of domain knowledge.

We identify several applications of ontologies in IE workflows to enhance LLM performance. Incorporating domain-specific ontologies during training could improve understanding of domain terms by supplying definitions, properties, and hierarchical relations, aligning the model's comprehension with that of domain experts. Ontologies can also guide semantic modeling by constraining or informing LLMs, improving interoperability and making outputs more actionable. For entity recognition (ER/NER) and relation extraction (RE), pre-informing models about common domain-specific terms and relationships may enhance accuracy by effectively teaching the model "what to look for." Additionally, retrieval-augmented generation (RAG) integrates knowledge bases, such as ontologies or knowledge graphs, into QA systems, reducing hallucinations and improving performance, as demonstrated in domains like biomedicine \cite{soman2024biomedical}.

Finally, these advantages need to be balanced with better understanding of what such constraints on LLMs might mean for their use in domains which have less well-structured general knowledge, perhaps because they are still emerging, or perhaps because the knowledge in those domains is changing rapidly. In such cases, there will be a need for further work to understand whether and to what extent using ontologies to constrain LLMs may also constrain our ability to track emerging or rapidly changing knowledge.

\section{Conclusion} 

This study demonstrates the potential of LLMs for advancing IE in invasion biology by mining key entities such as species, locations, habitats, and ecosystems from scientific literature. By developing and applying a standardized semantic schema, we showcased the ability of LLMs to structure complex ecological data, providing a foundation for enhancing workflows in ecological research. Using a two-stage approach, the specialize stage extracts detailed, context-specific structures, while the generalize stage integrates these into a flexible schema balancing specificity and generality. This method addresses the complexity of ecological systems, enabling structured representation of nuanced information. The release of the dataset and schema enable refining extraction methods, integrating ontologies, and exploring broader ecological applications. This research underscores the utility of LLMs as tools for bridging unstructured data and structured knowledge in ecology.

\bibliographystyle{acl_natbib}
\bibliography{nodalida2025}

\end{document}